\documentclass[letterpaper]{article}
\usepackage{aaai17}
\usepackage{times}
\pdfoutput=1
\usepackage{graphicx}
\usepackage{url}
\usepackage{graphicx}
\usepackage{amsmath}
\usepackage{amsfonts}
\usepackage{todonotes}
\usepackage{epstopdf}
\usepackage{color}
\usepackage{verbatim}
\usepackage{subfigure}
\title{Deep Learning for Unsupervised Insider Threat Detection\\in Structured Cybersecurity Data Streams}
\author{Aaron Tuor \and Samuel Kaplan \and Brian Hutchinson\thanks{Email: Brian.Hutchinson@wwu.edu.  Phone: 360-650-4894.  Address: 516 High Street, Bellingham, WA 98229.}\\
Western Washington University\\
 Bellingham, WA% 98225\\
 \AND Nicole Nichols \and Sean Robinson \\ Pacific Northwest National Laboratory\\ Seattle, WA}
\begin{document}
    \maketitle
     \begin{abstract}
    Analysis of an organization's computer network activity is a key component of early detection and mitigation of insider threat, a growing concern for many organizations. 
    Raw system logs are a prototypical example of 
    streaming data that can quickly scale beyond the cognitive power of a human analyst. 
    As a prospective filter for the human analyst, we present an online unsupervised deep learning approach to detect anomalous network activity from system logs in real time. 
    Our models decompose anomaly scores into the contributions of individual user behavior features for increased interpretability to aid analysts reviewing potential cases of insider threat.
     Using the CERT Insider Threat Dataset v6.2 and threat detection recall as our performance metric, our novel deep and recurrent neural network models outperform Principal Component Analysis, Support Vector Machine and Isolation Forest based anomaly detection baselines. 
       For our best model, the events labeled as insider threat activity in our dataset had an average anomaly score in the 95.53 percentile, demonstrating our approach's potential to greatly reduce analyst workloads.  
     \end{abstract}

 %----------------------------------------------------------------------------------------
 %	INTRODUCTION
 %----------------------------------------------------------------------------------------
 \section{Introduction}
 %\subsubsection{Insider Threat}
Insider threat is a complex and growing challenge for employers.
It is generally defined as any actions taken by an employee which are potentially harmful to the organization; e.g., unsanctioned data transfer or sabotage of resources.  
Insider threat may manifest in various and novel forms motivated by differing goals, ranging from a disgruntled employee subverting the prestige of an employer to advanced persistent threats (APT), orchestrated multi-year campaigns to access and retrieve intelligence data \cite{hutchins2011intelligence}.

 Cyber defenders 
 are tasked with assessing a large volume of real-time data. These datasets are high velocity, heterogeneous streams 
 generated by a large set of possible entities (workstations, servers, routers) and activities (DNS requests, logons, file accesses).
  With the goal of efficient utilization of human resources, automated methods for filtering system log data for an analyst have been the focus of much past and current  research, this work included.    

We present an online unsupervised deep learning system to filter system log data for analyst review. 
Because insider threat behavior is widely varying, we do not attempt to explicitly model threat behavior.
Instead, novel variants of Deep Neural Networks (DNNs) and Recurrent Neural Networks (RNNs) are trained to recognize activity that is characteristic of each user on a network and concurrently 
assess whether user behavior is normal or anomalous, all in real time.   
With the streaming scenario in mind, the time and space complexity of our methods are constant as a function of stream duration; that is, no data is cached indefinitely and detections are made as rapidly as new data is fed into our DNN and RNN models.
To aid analysts in interpreting system decisions, our model decomposes anomaly scores into a human readable summary of the major factors contributing to the detected anomaly (e.g. that the user copied an abnormally large number of files to removable media between 12am and 6am). 

There are several key difficulties in applying machine learning to the cyber security domain \cite{Sommer2010outside} that our model attempts to address. 
User activity on a network is often unpredictable over seconds to hours and contributes to the difficulty in finding a stable model of ``normal'' behavior. 
Our model trains continuously in an online fashion to adapt to changing patterns in the data. 
Also, anomaly detection for malicious events is particularly challenging because attackers often try to closely mimic typical behavior. 
We model the stream of system logs as interleaved user sequences with user-metadata to provide precise context for activity on the network; this allows our model, for example, to identify what is truly typical behavior for the user, employees in the same role, employees on the same project team, etc.
We assess the effectiveness of our models on the synthetic CERT Insider Threat v6.2 dataset \cite{lindauer2014generating,glasser2013bridging} which includes system logs with line-level annotations of insider threat activity.  The ground truth threat labels are used only for evaluation.

%----------------------------------------------------------------------------------------
%	PRIOR WORK
%----------------------------------------------------------------------------------------
\section{Prior Work}

A frequent approach to insider threat detection is to frame the problem as an anomaly detection task.  
A comprehensive overview of anomaly detection provided by Chandola et al. \shortcite{chandola2012anomaly} concludes that anomaly detection techniques for online and multivariate sequences are underdeveloped; both issues are addressed in this paper.
A real world system for anomaly detection in system logs should address the set of constraints given by the real time nature of the task and provide a set of features suitable for the application domain: concurrent tracking of multiple entities, analysis of structured multivariate data, adaptation to shifting distribution of activities, and interpretable judgments. 
While each work surveyed below addresses some subset of these components, our work addresses all of these constraints and features. 

As mentioned above, it is common to approach tasks like intrusion detection or insider threat as anomaly detection.  
Carter and Streilein \shortcite{carter2012probabilistic} demonstrate a probabilistic extension of an exponentially weighted moving average for the application of anomaly detection in a streaming environment. This method learns a parametric statistical model that adapts to the changing distribution of streaming data. An advantage of our present approach using deep learning architectures is the ability to model a wider range of distributions with fewer underlying assumptions.
Gavai et al. \shortcite{gavai2015supervised} compare a supervised approach, from an expert-developed classifier, with an unsupervised approach using the Isolation Forest method at the task of detecting insider threat from network logs. 
They also aggregate information about which features contribute to the isolation of a point within the tree to produce motivation for why a user was flagged as anomalous. 
Considering this to be a reasonable approach,
we include Isolation Forests as one of our baselines. 

Researchers have also applied neural network-based approaches to cybersecurity tasks.
Ryan et al. \shortcite{ryan1998intrusion} train a standard neural network with one hidden layer 
to predict  the probabilities that each of a set of ten users created a distribution of Unix commands for a given day. They detect a network intrusion when the probability is less than 0.5 for all ten users of the network. Differing from our work, their input features are not structured, and they do not train the network in an online fashion.
Early work on modeling normal user activity on a network using RNNs was performed by Debar  et al. \shortcite{debar1992neural}. They train an RNN to convergence on a representative sequence of Unix command line arguments (from login to logout) and predict network intrusion when the trained network for that user does poorly at predicting the login to logout sequence. While this work partially addresses online training it does not continuously train the network to take into account changing user habits over time.
Veeramachananeni et al.  \shortcite{veeramachaneniai2} present work using a neural network auto-encoder in an online setting. They aggregate numeric features over a time window from web and firewall logs which are fed to an ensemble of unsupervised anomaly detection methods: principal component reconstruction of the signal, auto-encoder neural network, and a multivariate probabilistic model over the feature space.
They additionally incorporate analyst feedback to continually improve with time, but do not explicitly model individual user activity over time.

Recurrent neural networks have, of course,
been successfully applied to anomaly detection in various alternative domains; e.g.,
Malhotra et al. \shortcite{malhotra2016lstm} in the domain of signals from mechanical sensors for machinery such as engines, and vehicles,  
Chuahan et al. \shortcite{chauhan2015anomaly} in the domain of ECG heart data, and
Marchi et al. \shortcite{marchi2015novel,marchi2015non} in the acoustic signal processing domain.
In contrast to the present work, these applications are not faced with the task of processing a multivariate combination of categorical and continuous features. 

\section{System Description}

\begin{figure}
    \centering
    \includegraphics[width=0.47\textwidth]{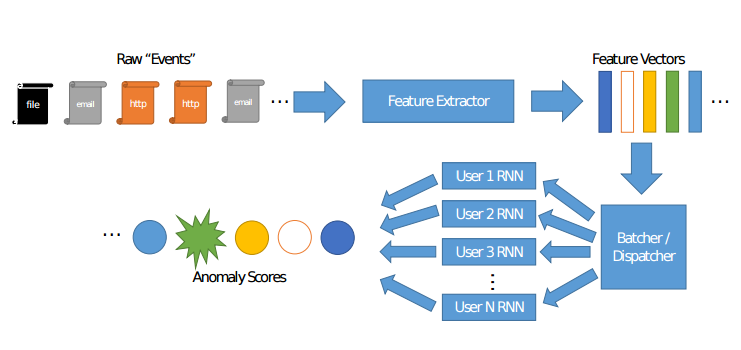}
    \caption{End to End System} \label{fig:system}
\end{figure}
Figure \ref{fig:system} provides an overview of our anomaly detection system.   First, raw events from system user logs are fed into our feature extraction system, which aggregates their counts and outputs one vector for each user for each day.  A user's feature vectors are then fed into a neural network, creating a set of networks, one per user.  In one variant of our system, these are DNNs; in the other, they are RNNs.  In either case, the different user models share parameters, but for the RNN they maintain separate hidden states.  These neural networks are tasked with predicting the next vector in the sequence; in effect, they learn to model users' ``normal'' behavior.  Anomaly is proportional to the prediction error, with sufficiently anomalous behavior being flagged for an analyst to investigate.  
The components in the system are described in greater detail below.

\subsection{Feature Extraction}

\begin{figure}
    \centering
    \includegraphics[width=0.47\textwidth]{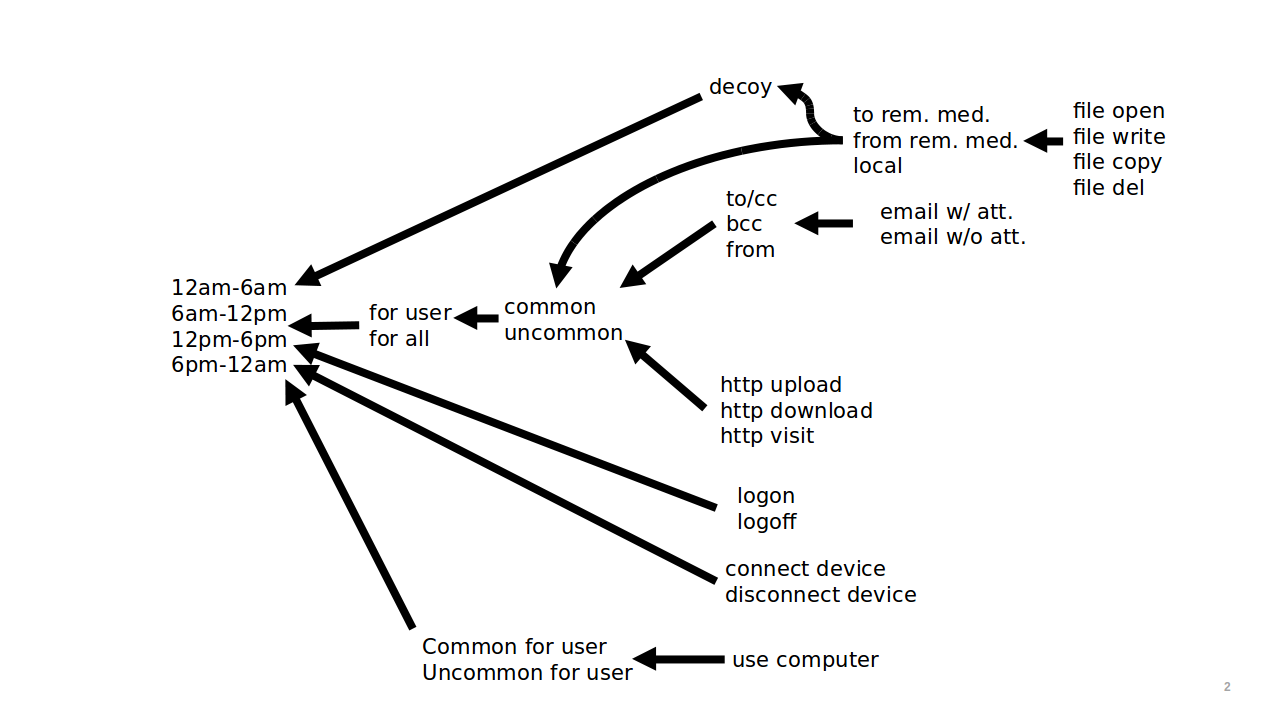}
    \caption{Enumeration of count features.} \label{fig:activities}
\end{figure}

One practical consideration that a deep learning anomaly detection system 
must address is the transformation of system log lines from heterogeneous tracking sources into numeric features suitable as input. 
Our system extracts two kinds of information from these sources: categorical user attribute features and continuous ``count'' features.  
The categorical user features refer to attributes such as a user's role, department, and supervisor in the organization.  
See Table \ref{tab:categoricalsizes} for a list of categorical features used in our experiments (along with the number of distinct values in each category). 
 In addition to these categorical features, we also accumulate counts of 408 ``activities'' a user has performed over some fixed time window (e.g. 24 hours).  
An example of a counted activity is the number of uncommon non-decoy file copies from removable media between the hours of 12:00 p.m. and 6:00 p.m.
Figure \ref{fig:activities} visually enumerates the set of count features: simply follow a path from right to left, choosing one item in each set along the way.  The set of all such traversals is the set of count features.
For each user $u$, for each time period, $t$, the categorical values and activity counts are concatenated into a 414 dimensional numeric feature vector ${\textbf x}_t^u$. 

\begin{table}
\centering
\begin{tabular}{|l|r|} \hline
    {\bf Categorical Var.} & {\bf \# Unique Values}\\\hline
    Role & 46\\
    Project & 366\\
    Functional Unit & 11\\
    Department & 23\\
    Team & 90\\
    Supervisor & 246 \\ \hline
\end{tabular}
\caption{Categorical Variables} \label{tab:categoricalsizes}
\end{table}

\subsection{Structured Stream Neural Network} \label{sec:ssnn}

At the core of our system is one of two neural network models that map a series of feature vectors for a given user, one per day, to a probability distribution over the next vector in the user's sequence.  This model is trained jointly over all users simultaneously and in an online fashion.  First, we describe our DNN model, which does not explicitly model any temporal behavior, followed by the RNN, which does.   We then discuss the remaining components for making predictions of structured feature vectors and identification of anomaly in the stream of feature vectors.

\subsubsection{Deep Neural Network Model}
Our model takes as input a series of $T$ feature vectors ${\bf x}_1^u,{\bf x}_2^u,\dots,{\bf x}_T^u$ for a user $u$ and produces as output a series of $T$ hidden state vectors ${\bf h}_1^u,{\bf h}_2^u,\dots,{\bf h}_T^u$ (each to be later fed into the structured prediction network).  In a DNN with $L$ hidden layers $(l=1,...,L)$, our final hidden state, the output of hidden layer $L$, ${\bf h}_t^u = {\bf h}_{L,t}^u$ is a function of ${\bf x}_t^u$ as follows: 

\begin{equation}
    {\bf h}_{l,t}^u  =  g({\bf W}_{l} {\bf h}_{l\text{-}1,t}^u + {\bf b}_l)
\end{equation}

Where $g$ is a non-linear activation function, typically ReLU, tanh, or the logistic sigmoid, and ${\bf h}_{0,t}^u = {\bf x_t^u}$. The trainable parameters are the $L$ weight matrices ($\bf{W}$), and $L$ bias vectors ($\bf{b}$).

\subsubsection{Recurrent Neural Network Model}

\begin{figure}
    \centering
    \includegraphics[width=0.47\textwidth]{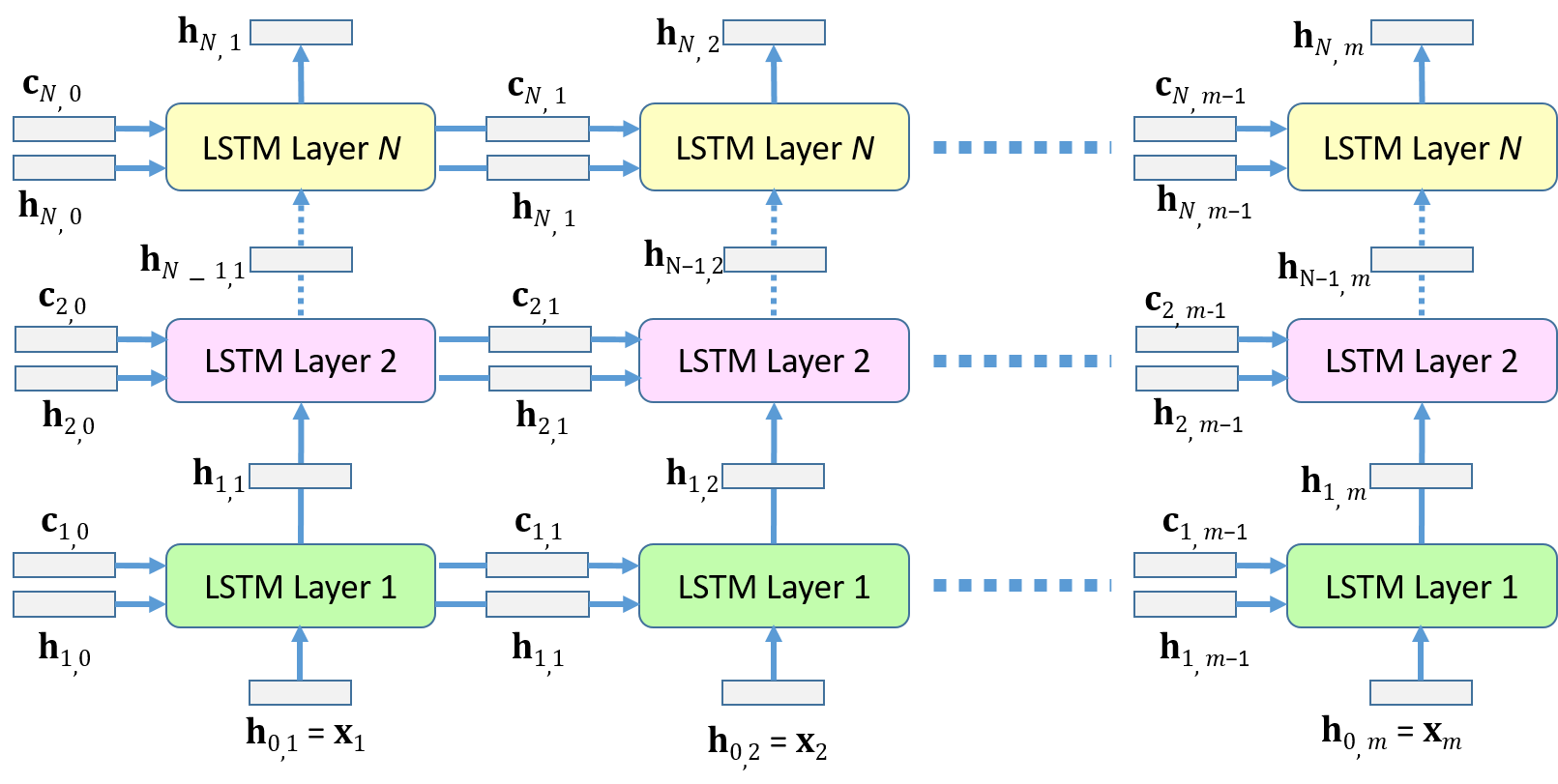}
    \caption{Unrolled LSTM Network with $N$ Layers} \label{fig:lstm}
\end{figure}

Like the DNN, the RNN model maps an input sequence ${\bf x}_1^u,{\bf x}_2^u,\dots,{\bf x}_t^u$ to a hidden state sequence ${\bf h}_1^u, {\bf h}_2^u, \dots, {\bf h}_T^u$.  Unlike the DNN, here the hidden state ${\bf h}_t^u$ is computed as a function of ${\bf x}_1^u,{\bf x}_2^u,\dots,{\bf x}_t^u$, and not on ${\bf x}_t^u$ alone.  Conditioning ${\bf h}_t^u$ on a sequence rather than the current input alone allows us to capture temporal patterns in user behavior, and to build an increasingly accurate model of the user's behavior over time.

We use the popular Long Short-Term Memory (LSTM) RNN architecture \cite{hochreiter1997long}, in which the hidden state ${\bf h}_t^u$ at time $t$ is a function of a long-term memory cell, ${\bf c}_t^{u}$.  In a deep LSTM with $L$ hidden layers, our final hidden state, the output of hidden layer $L$, ${\bf h}_t^u = {\bf h}_{L,t}^u$, depends on the input sequence and cell states as follows: 
\begin{eqnarray}
    {\bf h}_{l,t}^u & = & {\bf o}_{l,t}^u \odot \tanh({\bf c}_{l,t}^u)\\
    {\bf c}_{l,t}^u & = & {\bf f}_{l,t}^u \odot {\bf c}_{l,t\text{-}1}^u + {\bf i}_{l,t}^u \odot {\bf g}_{l,t}^u \mbox{, and }\\
    {\bf g}_{l,t}^u & = & \tanh\left( {\bf W}_{l}^{(g,x)} {\bf h}_{l\text{-}1,t}^u + {\bf W}_{l}^{(g ,h)}{\bf h}_{l,t\text{-}1}^u + {\bf b}_l^g \right)\\
    {\bf f}_{l,t}^u & = & \sigma\left( {\bf W}_{l}^{(f,x)} {\bf h}_{l\text{-}1,t}^u + {\bf W}_{l}^{(f ,h)}{\bf h}_{l,t\text{-}1}^u + {\bf b}_l^f \right)\\
   {\bf i}_{l,t}^u & = & \sigma\left( {\bf W}_{l}^{(i,x)} {\bf h}_{l\text{-}1,t}^u + {\bf W}_{l}^{(i ,h)}{\bf h}_{l,t\text{-}1}^u + {\bf b}_l^i \right)\\
     {\bf o}_{l,t}^u & = & \sigma\left( {\bf W}_{l}^{(o,x)} {\bf h}_{l\text{-}1,t}^u + {\bf W}_{l}^{(o ,h)}{\bf h}_{l,t\text{-}1}^u + {\bf b}_l^o \right)
\end{eqnarray}
Where ${\bf h}_{0,t}^u = {\bf x_t^u}$, and ${\bf c}_{l,0}^u$, ${\bf h}_{l,0}^u$ are set to zero vectors for all $1\leq l \leq L$.
We use $\odot$ and $\sigma$ to denote element-wise multiplication and the (element-wise) logistic sigmoid function, respectively.  Vector ${\bf g}_{l,t}^u$ is a hidden representation based on the current input and previous hidden state, while vectors ${\bf f}_{l,t}^u$, ${\bf i}_{l,t}^u$ and ${\bf o}_{l,t}^u$, modulate how cell-state information is propagated across time, how the input is incorporated into the cell state, and how the the hidden state relates to the cell state, respectively.  The trainable parameters for the LSTM are the $8L$ weight matrices (${\bf W}$) and the $4L$ bias vectors (${\bf b}$); these weights are shared among all users.

\subsubsection{Probability Decomposition} 
Given the hidden state at time $t-1$, ${\bf h}_{t-1}^{u}$, our model outputs the parameters $\theta$ for a probability distribution over the next observation, ${\bf x}_t^{u}$.  The anomaly for user $u$ at time $t$, $a_t^u$, is then:
\begin{equation}
    a_t^u = - \log P_\theta({\bf x}_t^{u}|{\bf h}_{t-1}^{u}) \label{eqn:nexttimestep}
\end{equation}
This probability is complicated by the fact that our feature vectors, and thus the predictions our model makes, include six categorical variables in addition to the 408 dimensional count vector.  Therefore, $P_\theta({\bf x}_t^{u}|{\bf h}_{t-1}^{u})$ is actually the joint probability over the count vector ($\hat{\bf x}_t^u$) and each of the categorical variables: role (R), project (P), functional unit (F), department (D), team (T) and supervisor (S).  Let ${\cal C} = \lbrace R, P, F, D, T, S \rbrace$ denote the set of categorical variables; e.g., let $R_t^u$ denote the role of user $u$ at time $t$.  Then
\begin{equation}
    P_\theta({\bf x}_t^{u}|{\bf h}_{t-1}^{u}) = P_\theta(\hat{\bf x}_t^u,R_t^{u},\dots,S_t^{u}|{\bf h}_{t-1}^{u}).
\end{equation}
For computational simplicity, we approximate this joint probability by assuming conditional independence:
\begin{equation}
    P_\theta({\bf x}_t^{u}|{\bf h}_{t-1}^{u}) \approx
    P_{\bf \theta^{(\hat{\bf x})}}(\hat{\bf x}_t^u|{\bf h}_{t-1}^{u})
    \prod_{V \in {\cal C}} P_{{\bf \theta}^{(V)}}(V_t^u|{\bf h}_{t-1}^{u})
\end{equation}
The seven parameter vectors, parameters ${\bf \theta^{(\hat{\bf x})}}$ and ${\bf \theta}^{(V)}$ for $V \in {\cal C}$, are produced by seven single hidden layer neural networks:
\begin{eqnarray}
    {\bf \theta}_t^{({\bf \hat{x}})} & = & {\bf U}'_{\bf \hat{x}} \tanh\left( {\bf U}_{\bf \hat{x}} {\bf h}_{t-1} + {\bf b}_{\bf \hat{x}}\right) + {\bf b}'_{\bf \hat{x}})\\
    {\bf \theta}_t^{(V)} & = & f({\bf U}'_V \tanh\left( {\bf U}_V {\bf h}_{t-1} + {\bf b}_V\right) + {\bf b}'_V)
\end{eqnarray}
Here $f$ denotes the softmax function.  Two additional weight matrices (${\bf U}$) and two additional bias vectors $({\bf b})$ are introduced for each of the seven variables we are predicting.  Like the LSTM weights, these parameters are shared among all users.  The parametric forms for the conditional probabilities are described next.

\subsubsection{Conditional Probabilities}
We model the conditional probabilities for the six categorical variables as discrete, while we model the conditional probability of the counts as continuous.  For the discrete models, we use the standard approach: the probability of category $k$ is simply the $k$th element of vector $\theta^{(V)}$, whose dimension is equal to the number of categories.  
For example, there are 47 roles, so $\theta^{(R)} \in \mathbb{R}^{47}$. 
Because we use a softmax output activation to produce $\theta^{(V)}$, the elements are non-negative and sum-to-one.

For the count vector, we use the multivariate normal density: $P_{\bf \theta^{(\hat{\bf x})}}(\hat{\bf x}_t^u|{\bf h}_{t-1}^{u}) = {\cal N}(\hat{\bf x};\mu,\Sigma)$.  
We consider two variants.  
In the first, our model outputs the mean vector $\mu$ ($\theta^{({\bf \hat{x}})}=\mu$) and we assume the covariance $\Sigma$ to be the identity.  
With identity covariance, maximizing the log-likelihood of the true data is equivalent to minimizing the squared error $\Vert \hat{\bf x}_t^u - {\bf \mu}\Vert^2$.  
In the second, we assume diagonal covariance, and our model outputs both the mean vector and the log of the diagonal of $\Sigma$. 
This portion of the model can be seen as a simplified Mixture Density Network \cite{bishop94}.

\subsubsection{Prediction Targets}
We define two prediction target approaches, ``next time step" and ``same time step". Recall from Eqn. \ref{eqn:nexttimestep}, anomaly is inversely proportional to the log probability of the observation at time $t$ given the hidden representation at time $t\text{-}1$; that is, given everything we know up to and including time $t\text{-}1$, predict the outcome at time $t$.  This approach fits the normal paradigm for RNNs on sequential data; in our experiments, we will refer to this approach as ``next time step'' prediction.

However, it is common in anomaly detection literature \cite{malhotra2016lstm} to use an auto-encoder to detect anomaly. 
 An auto-encoder is a parametric function trained to reproduce the input features as output.  
 Its complexity is typically constrained to prevent it from learning the trivial identity function; instead, the network must exploit statistical regularities in the data to achieve low reconstruction error for commonly found patterns, at the expense of high reconstruction error for uncommon patterns (anomalous activity).  Networks trained in this unsupervised fashion have been demonstrated to be very effective in several anomaly detection application domains \cite{markou2003novelty}.

In the context of our present application, both techniques may be applicable.  Formally, we consider an alternative definition of anomaly:
\begin{equation}
    \hat{a}_t^u = - \log P_\theta({\bf x}_t^{u}|{\bf h}_t^{u}) \label{eqn:sametimestep}
\end{equation}
That is, given everything we know up to and including time $t$, predict the input counts ${\bf x}_t^u$.  If ${\bf x}_t^u$ is anomalous, we are unlikely to produce a distribution that assigns a large density to it.  We refer to this approach as ``same time step'' prediction.

\subsubsection{Detecting Insider Threat}
Ultimately, the goal of our model is to detect insider threat.  We assume the following conditions: our model produces anomaly scores, which are used to rank user-days from most anomalous to least, we then provide the highest ranked user-day pairs to analysts who judge whether the anomalous behavior is indicative of insider threat.  We assume that there is a daily budget which imposes a maximum number of user-day pairs that can be judged each day, and that if an actual case of insider threat is presented to an analyst, he or she will correctly detect it. 

Because our model is trained in an online fashion, the anomaly scores start out quite large (when the model knows nothing about normal behavior) and trend lower over time (as normal behavior patterns are learned). To place the anomaly score for user $u$ at time $t$ in the proper context, we compute an exponentially weighted moving average estimate of the mean and variance of these anomaly scores and standardize each score as it arrives.

One key feature of our model is that the anomaly score decomposes as the sum over the negative log probabilities of our variables; the continuous count random variable further decomposes over the sum of individual feature terms: $(x_i-\mu_i)/\sigma_i$.  This allows us to identify which features are largest contributors to any anomaly score; for example, our model could indicate that a particular user-day is flagged as anomalous primarily due to an abnormal number of emails sent with attachments to uncommon recipients between 12am and 6am.  Providing insight into {\it why} a user-day was flagged may improve both the speed and accuracy of analysts' judgments about insider threat behavior.

\subsection{Online Training}
In a standard training scenario for RNNs, individual or mini-batches of sequences are fed to the RNN, gradients of the training objective are computed via Back Propagation Through Time, and then weights are adjusted via a gradient-descent-like algorithm. 
For DNNs, individual or mini-batches of samples are fed into the DNN, and weights are updated with gradients computed by standard backpropagation.  
In either case, this process usually iterates over the fixed-size dataset until the model converges, and only then is the model applied to new data to make predictions. 
This approach faces a few key challenges for the online anomaly detection setting: 1) the dataset is streaming and effectively unbounded and 2) the model is tasked with making predictions on new data as it learns.  
Attempting to shoehorn this scenario into a standard training setup is impractical: it is infeasible to either store or repeatedly to train on an unbounded streaming dataset and periodically retraining the model on a fixed-size set of recent events risks excluding important past events.  

To accommodate 
an online 
scenario, we make important adjustments to the standard training regimen. 
For DNNs, the primary difference is the restriction of observing each sample only once.
For the RNN, the situation is more complicated.  
We train on multiple user sequences concurrently, backpropagating and adjusting weights each time we see a new feature vector from a user.  
Logically, this corresponds to training one RNN per user, where the weights are shared between all users but hidden state sequences are per-user.  
In practice, we accomplish this by training a single RNN with a supplementary data structure that stores a finite window of past inputs and hidden and cell states for each user. 
Each time a new feature vector for a user is fed into the model, the hidden and cell states for that user are then used for context when calculating the forward pass and backpropagating error. 

\subsection{Baseline Models}

To assess the effectiveness of our DNN and RNN models, we compare against popular anomaly/novelty/outlier detection methods. 
Specifically, we compare against one-class support vector machine (SVM) \cite{Scholkopfetal01}, isolation forest \cite{LiuTingZhou08} and principle component analysis (PCA) baselines \cite{ShyuChenetal03}.  We use scikit-learn's\footnote{http://scikit-learn.org/stable/modules/outlier\_detection.html} implementation of one-class SVM and isolation forest, both included as part of its novelty and outlier detection functionality \cite{scikit-learn}.
For the PCA baseline, we project the feature vector onto the first $k$ principle components and then map it back into the original feature space.  Anomaly is proportional to the error in this reconstruction.  Hyperparameter $k$ is tuned on the development set. 
 
%----------------------------------------------------------------------------------------
%	EXPERIMENTS
%----------------------------------------------------------------------------------------
\section{Experiments}
%\cite{tensorflow2015-whitepaper},
We assess the effectiveness of our model, which we implemented in Tensorflow\footnote{Code will be available at https://github.com/pnnl/safekit} \cite{tensorflow2015-whitepaper}
 on a series of experiments.  In this section we describe the data used, hyper-parameters tuned, and present our results and analysis.

\subsection{Data} 

Given security and privacy concerns surrounding network data, real world datasets must undergo an anonymization process before being publicly released for research purposes. 
The anonymization process may obscure potentially relevant factors in system logs. 
Particularly, user attribute metadata that may be available to a system administrator is typically absent in an open release data set.  
We perform experiments on the synthetic CERT Insider Threat Dataset v6.2, which includes such categorical information.
 
CERT consists of event log lines from a simulated organization's computer network, generated with sophisticated user models.
We use five sources of events: logon/logoff activity, http traffic, email traffic, file operations, and external storage device usage.  
Over the course of 516 days, 4,000 users generate 135,117,169 events (log lines).
Among these are events manually injected by domain experts, representing five insider threat scenarios taking place.  
Additionally, user attribute metadata is included; namely, the six categorical attributes listed in Table \ref{tab:categoricalsizes}.

Since this is an unsupervised task, no supervised training set is required.
We therefore split the entire dataset chronologically into two subsets: development and test. 
The former subset ($\sim$85\% of the data) is used for model selection and hyper-parameter tuning, while the latter subset ($\sim$15\% of the data) is held out for assessing generalization performance.
Table \ref{tab:datasplit} summarizes the dataset statistics.  
Our predictions are made at the granularity of user-day; there are fewer threat user-days than raw events because malicious users often conduct several threat events over the course of a single day.  
Note that although the test set includes only 15\% of the events, it has over 40\% of the threat user-days.  One final note is that we filtered our data to keep only weekdays, because what is normal is qualitatively different for weekdays and weekends.  If desired, a second system could be trained to model normal weekend behavior.

\begin{table}
    \centering
    \begin{tabular}{l|r|r|}
        \cline{2-3}
                                               & {\bf Development}   & {\bf Test}        \\ \hline
        \multicolumn{1}{|l|}{\it Date Range}       & {\it Days 1 - 418} & {\it Days 419 - 516} \\ \hline
        \multicolumn{1}{|l|}{\# Device Events} & 1,285,341            & 266,487             \\ \hline
        \multicolumn{1}{|l|}{\# Email Events}  & 9,068,429            & 1,926,528           \\ \hline
        \multicolumn{1}{|l|}{\# File Events}   & 1,671,698            & 343,185             \\ \hline
        \multicolumn{1}{|l|}{\# HTTP Events}   & 96,516,038           & 20,509,178          \\ \hline
        \multicolumn{1}{|l|}{\# Logon Events}  & 2,916,161            & 614,124             \\ \hline
        \multicolumn{1}{|l|}{\bf Total Events} & {\bf 111,457,667}    & {\bf 23,659,502}    \\ \hline\hline
        \multicolumn{1}{|l|}{\textcolor{red}{Threat Events}}    & \textcolor{red}{192}                  & \textcolor{red}{236}     \\ \hline
        \multicolumn{1}{|l|}{\textcolor{red}{Threat User-Days}} & \textcolor{red}{27}                   & \textcolor{red}{20}\\ \hline
    \end{tabular}
    \caption{Dataset statistics.} \label{tab:datasplit}
\end{table}

\subsection{Tuning}
We tune our models and baselines on the development set using random hyper-parameter search. 
For the DNN, we tune the number of hidden layers (between 1 and 6) and the hidden layer dimension (between 20 and 500). 
We fix the batch size to 256 samples (user-days) and the learning rate to 0.01. 
For the RNN, we tune the hidden layers and hidden layer dimension over the same ranges as the DNN, and also fix the learning rate to 0.01. The batch size is tuned (between 256 and 8092 samples); larger batch sizes speed up model training, which is more important for the RNN than the DNN.  We also tune the number of time steps to back propagate over (between 3 and 40). When our inputs and outputs include the categorical variables, we additionally tune a hyper-parameter which determines the size of the input embedding vector of a category in relation to how many classes in that category (between 0.25 and 1).  Both neural network models use $\tanh$ for the hidden activation function and are trained using the ADAM \cite{kingma2014adam} variant of gradient descent. 

We also tune our baseline models.  
For the PCA baseline, we tune over the number of principal components (between 1 and 20).  
For the Isolation Forest baseline, we tune the number of estimators (between 20 and 300), the contamination (between 0 and 0.5), and whether we bootstrap (true or false).  
The max feature hyper-parameter is fixed at the default of 1.0 (use all features).  
For the SVM baseline, we tune the kernel (in the set \{\verb|rbf|, \verb|linear|, \verb|poly|, \verb|sigmoid|\}),   $\nu$ (between 0 and 1) and whether to use the shrinking heuristic (true or false).  
For the polynomial kernel, we tune the degree (between 1 and 10) while for all other kernels we use the default value for the remaining hyper-parameters.

For all models, our tuning criteria is Cumulative Recall $k$ (CR-$k$), which we define to be the sum of the recalls for all budgets up to and including $k$.  
For computational efficiency, we only evaluate budgets at increments of 25, so if we defined $R(i)$ to be the recall with a budget of $i$, CR-$k$ is actually $R(25)+R(50)+\cdots+R(k)$.  
CR-$k$ can be thought of as an approximation to an area under the recall curve. For each model, we picked the hyper-parameters that maximized CR-1000, for which the maximum value achievable is 40.
Given the assumptions that 1) we have a fixed daily analyst budget which cannot be carried over from one day to the next, 2) true positives are rare, and 3) the cost of a missed detection is substantially larger than the cost of a false positive, we feel that recall-oriented metrics such as CR-$k$ are a more suitable measurement of performance than precision-oriented ones.

\subsection{Results}
We present three sets of experimental results, each designed to answer a specific question about our model's performance.  

\begin{table}
    \centering
    \begin{tabular}{|l||r|r|}\hline
        {\bf Model}  & {\bf CR-400} & {\bf CR-1000} \\\hline\hline
        {\bf LSTM-Diag} & 11.6 & 35.6 \\\hline
        {\bf LSTM-Diag-Cat} & 9.2 & 32.3 \\\hline
    \end{tabular}
    \caption{Cumulative Recall (CR-$k$) for budgets of 400 and 1000.  Comparing the performance of diagonal covariance LSTM models with (Cat) and without categorical features included.} \label{tab:categoricalcomparison}
\end{table}

First, we 
assess the effect of including or excluding the categorical variables in our model input and output.  
Table \ref{tab:categoricalcomparison} shows the comparison between two LSTM models, differing only in whether they include or exclude the categorical information.  
It shows that while the difference is not huge, the model clearly performs better without the categorical information.  
While the original intention of including categorical features was to provide context to the model, we hypothesize that our dataset may be simple enough that such context is not necessary (or that the model does not need explicit context: it can infer it).  
It may also be that the added model complexity hinders trainability, leading to a net loss in performance.  
Because inclusion of categorical features adds computational complexity to the model and harms performance, all of the remaining experiments reported in this paper use count features only.

\begin{table}
    \centering
    \begin{tabular}{|l||r|r|}\hline
        {\bf Model}  & {\bf CR-400} & {\bf CR-1000} \\\hline\hline
        {\bf LSTM-Diag} & 11.6 & 35.6 \\\hline
        {\bf LSTM-Diag-NextTime} & 5.9 & 25.1 \\\hline
        {\bf DNN-Diag} & 11.7 & 35.7 \\\hline
        {\bf DNN-Diag-NextTime} & 9.4 & 32.5 \\\hline
    \end{tabular}
    \caption{Cumulative Recall (CR-$k$) for daily budgets of 400 and 1000.  Comparing the performance of the diagonal covariance DNN and LSTM models predicting counts at the next time steps (NextTime) vs the current time step.} \label{tab:timestepcomparison}
\end{table}

Our second set of experiments is designed to determine which of the prediction modes work best for our task: ``same time step'' (Eqn. \ref{eqn:sametimestep}) or ``next time step'' (Eqn. \ref{eqn:nexttimestep}).  
Table \ref{tab:timestepcomparison} shows these results, comparing two DNN and two LSTM models.  
The ``same time step'' approach yields better performance for both models, although the difference is more dramatic for the LSTM.  Based on this result, we only use ``same time step'' for our remaining set of experiments.  Interestingly, the DNN and LSTM perform equivalently.  
We suspect that the CERT dataset does not contain enough temporal patterns unfolding over multiple days to offer any real advantage to the LSTM, though we would expect it to offer advantages on real-world datasets.

\begin{table}
    \centering
    \begin{tabular}{|l||r|r|}\hline
        {\bf Model}  & {\bf CR-400} & {\bf CR-1000} \\\hline\hline
        {\bf Isolation Forest} & 10.8 & 34.8 \\\hline 
        {\bf SVM} & 5.3 & 24.2 \\\hline
        {\bf PCA} & 9.4 & 32.8 \\\hline
        {\bf DNN-Ident} & 9.8 & 32.4 \\\hline
        {\bf DNN-Diag} & 11.7 & 35.7 \\\hline
        {\bf LSTM-Ident} & 10.8 & 33.0 \\\hline
        {\bf LSTM-Diag} & 11.6 & 35.6 \\\hline
    \end{tabular}
    \caption{Cumulative Recall (CR-$k$) for daily budgets of 400 and 1000.  All results are based on count features only.  For the DNN and LSTM, diagonal (Diag) and identity (Ident) covariances are contrasted.} \label{tab:maindata}
\end{table}

Our final set of experiments is designed to assess the effect of covariance type for our continuous features (identity versus diagonal) and to contrast with our baseline models.  Table \ref{tab:maindata} shows these results.  
Among the baselines, the Isolation Forest model is the strongest, giving the third best performance after DNN-Diag and LSTM-Diag.
These results also show that diagonal covariance leads to better performance than identity covariance.  One obvious advantage of diagonal covariance is that it is capable of more effectively normalizing the data (by accounting for trends in variance).  Wondering how well the identity model would perform if the data was normalized ahead of time, we conducted a pilot study where the counts were standardized with an exponentially weighted moving average estimate of the mean and variance, and found no improvement for either the identity or diagonal covariance models.  In contrast to a ``global'' normalization scheme, our diagonal covariance model is capable of conditioning the mean and variance on local context (when either ``next time step'' or the LSTM are used); for example, it might expect greater mean or variance in the number of emails sent on the day after an abnormally large number of emails were received.  That said, it is not clear whether our data exhibits patterns that our models can take advantage of with this dynamic normalization.

\subsection{Analysis}

\begin{figure}
    \centering
        \includegraphics[scale=0.44]{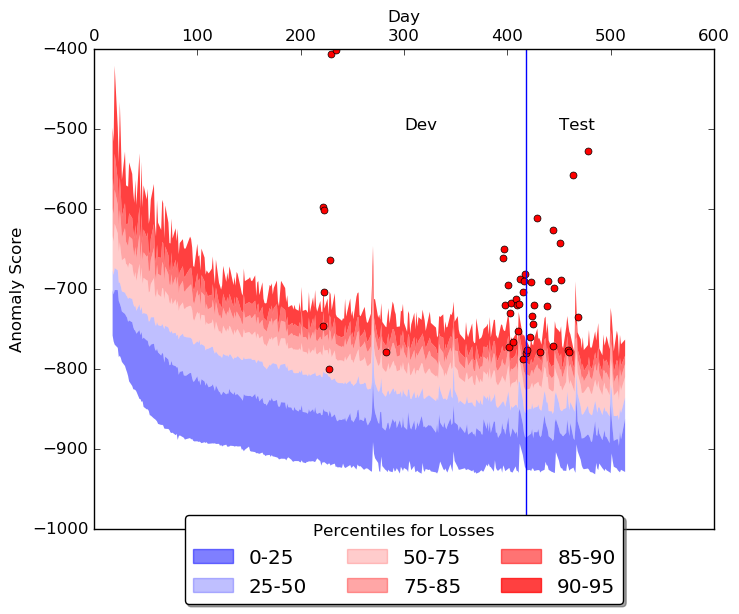}
        \caption{Percentile ranges of user-day anomaly as a function of days for the DNN-Diag model. The vertical bar denotes the split between the development and test sets.} \label{fig:percentiles}
\end{figure}

We perform two analyses to better understand our system's behavior, using our best DNN model to illustrate. 
In the first, we look at the effect of time on the model's notion of anomaly.  
Because the model begins completely untrained, anomaly scores for all users are very high for the first few days.  As the model sees examples of user behavior, it quickly learns what is ``normal.'' Fig. \ref{fig:percentiles} shows anomaly as a function of day, (starting after the ``burn in'' period of the first few days, to keep the y-axis scale manageable).  Percentile ranges are shown (computed over the users in the day), and malicious (insider threat) user-days are overlayed as red dots.  
Notice that all malicious events are above the 50th percentile for anomaly, with most above the 95th percentile.

\begin{figure}
    \centering
    \includegraphics[scale=0.43]{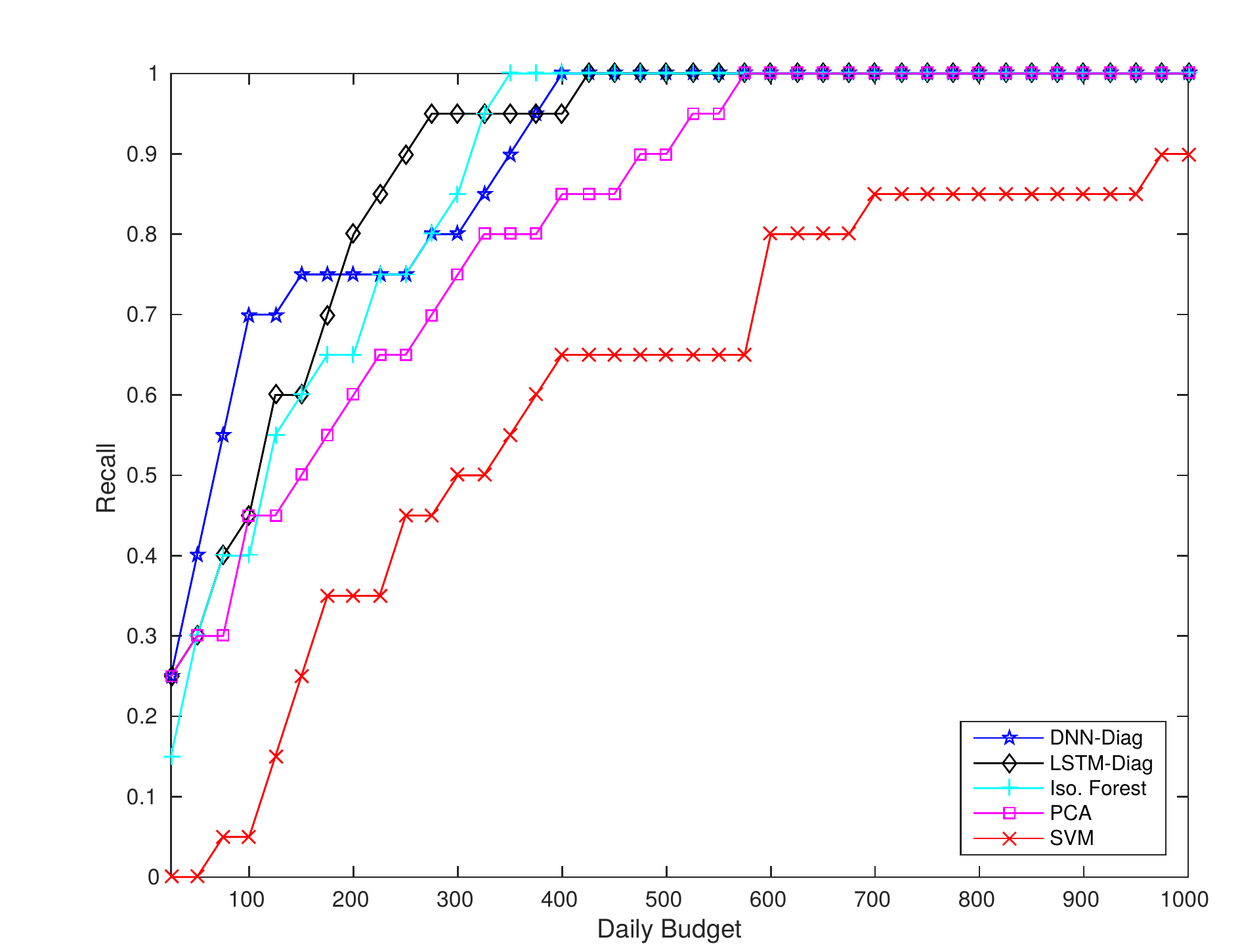}
    \caption{Test set recall curves.} \label{fig:auc_plot}
\end{figure}

In our second analysis, we study the effect of daily budget on recall for best DNN, best LSTM and the three baseline models.  Fig. \ref{fig:auc_plot} plots these recall curves.  Impressively, with a daily budget of 425, DNN-Diag, LSTM-Diag and the Isolation Forest model all obtain 100\% recall.  
It also shows that with our LSTM-Diag system, 90\% recall can be obtained with a budget of only 250 (a 93.5\% reduction in the amount of data analysts need to consider).

%----------------------------------------------------------------------------------------
%	CONCLUSIONS
%----------------------------------------------------------------------------------------
\section{Conclusions}
We have presented a system employing an online deep learning architecture that produces interpretable assessments of anomaly for the task of insider threat detection in streaming system user logs. Because insider threat takes new and different forms, it is not practical to explicitly model it; our system instead models ``normal'' behavior and uses anomaly as an indicator of potential malicious behavior.  Our approach is designed to support the streaming scenario, allowing high volume streams to be filtered down to a manageable number of events for analysts to  review.  Further, our probabilistic anomaly scores also allow our system to convey {\it why} it felt a given user was anomalous on a given day (e.g. because the user had an abnormal number of file uploads between 6pm and 12am).  We hope that this interpretability will improve human analysts' speed and accuracy.

In our evaluation using the CERT Insider Threat v6.2 dataset, our DNN and LSTM models outperformed three standard anomaly detection technique baselines (based on Isolation Forest, SVMs and PCA).  
When our probabilistic output model uses a context-dependent diagonal covariance matrix (as a function of the input) rather than a fixed identity covariance matrix, it provides better performance.  We also contrasted two prediction scenarios: 1) probabilistically reconstructing the current input given a compressed hidden representation (``same time step'') and 2) probabilistically predicting the next time step (``next time step'').  In our experiments, we found that the first works slightly better.

There are many ways one could extend this work.  First, we would like to apply this to a wider range of streaming tasks.  Although our focus here is on insider threat, our underlying model offers a domain agnostic approach to anomaly detection.  In our experiments, the LSTM performed equivalently to the DNN, but we suspect that the LSTM will yield superior performance when applied to large-scale real-world problems with more complicated temporal patterns. 

Another promising angle is to explore different granularities of times.  The current work aggregates features over individual users for each day; this has the potential to miss anomalous patterns happening within a single day.  Again, our LSTM model has the greatest potential to generalize: the model could be applied to individual events / log-lines, using its hidden state as memory to detect anomalous sequences of actions.  Doing so would reduce or eliminate the ``feature engineering'' required for aggregate count-style features. It could also dramatically narrow the set of individual events an analyst must inspect to determine whether anomalous behavior constitutes insider threat.

%----------------------------------------------------------------------------------------
%	ACKNOWLEDGMENTS
%----------------------------------------------------------------------------------------
\section{Acknowledgments.}
The research described in this paper is part of the Analysis in Motion Initiative at Pacific Northwest National Laboratory. It was conducted under the Laboratory Directed Research and Development Program at PNNL, a multi-program national laboratory operated by Battelle for the U.S. Department of Energy, and supported in part by the U.S. Department of Energy, Office of Science, Office of Workforce Development for Teachers and Scientists (WDTS) under the Visiting Faculty Program (VFP).

\bibliography{anomaly}

\begin{thebibliography}{}

\bibitem[\protect\citeauthoryear{Abadi \bgroup et al\mbox.\egroup
  }{2015}]{tensorflow2015-whitepaper}
Abadi, M.; Agarwal, A.; Barham, P.; Brevdo, E.; Chen, Z.; Citro, C.; Corrado,
  G.~S.; Davis, A.; Dean, J.; Devin, M.; Ghemawat, S.; Goodfellow, I.; Harp,
  A.; Irving, G.; Isard, M.; Jia, Y.; Jozefowicz, R.; Kaiser, L.; Kudlur, M.;
  Levenberg, J.; Man\'{e}, D.; Monga, R.; Moore, S.; Murray, D.; Olah, C.;
  Schuster, M.; Shlens, J.; Steiner, B.; Sutskever, I.; Talwar, K.; Tucker, P.;
  Vanhoucke, V.; Vasudevan, V.; Vi\'{e}gas, F.; Vinyals, O.; Warden, P.;
  Wattenberg, M.; Wicke, M.; Yu, Y.; and Zheng, X.
\newblock 2015.
\newblock {TensorFlow}: Large-scale machine learning on heterogeneous systems.
\newblock Software available from tensorflow.org.

\bibitem[\protect\citeauthoryear{Bishop}{1994}]{bishop94}
Bishop, C.
\newblock 1994.
\newblock Mixture density networks.
\newblock Technical Report NCRG/94/004, Neural Computing Research Group, Aston
  University.

\bibitem[\protect\citeauthoryear{Carter and
  Streilein}{2012}]{carter2012probabilistic}
Carter, K.~M., and Streilein, W.~W.
\newblock 2012.
\newblock Probabilistic reasoning for streaming anomaly detection.
\newblock In {\em Proc. SSP},  377--380.

\bibitem[\protect\citeauthoryear{Chandola, Banerjee, and
  Kumar}{2012}]{chandola2012anomaly}
Chandola, V.; Banerjee, A.; and Kumar, V.
\newblock 2012.
\newblock Anomaly detection for discrete sequences: A survey.
\newblock {\em IEEE TKDE} 24(5):823--839.

\bibitem[\protect\citeauthoryear{Chauhan and Vig}{2015}]{chauhan2015anomaly}
Chauhan, S., and Vig, L.
\newblock 2015.
\newblock Anomaly detection in ecg time signals via deep long short-term memory
  networks.
\newblock In {\em Proc. DSAA},  1--7.

\bibitem[\protect\citeauthoryear{Debar, Becker, and
  Siboni}{1992}]{debar1992neural}
Debar, H.; Becker, M.; and Siboni, D.
\newblock 1992.
\newblock A neural network component for an intrusion detection system.
\newblock In {\em Proc. IEEE Symposium on Research in Security and Privacy},
  240--250.

\bibitem[\protect\citeauthoryear{Gavai \bgroup et al\mbox.\egroup
  }{2015}]{gavai2015supervised}
Gavai, G.; Sricharan, K.; Gunning, D.; Hanley, J.; Singhal, M.; and Rolleston,
  R.
\newblock 2015.
\newblock Supervised and unsupervised methods to detect insider threat from
  enterprise social and online activity data.
\newblock {\em Journal of Wireless Mobile Networks, Ubiquitous Computing, and
  Dependable Applications} 6(4):47--63.

\bibitem[\protect\citeauthoryear{Glasser and
  Lindauer}{2013}]{glasser2013bridging}
Glasser, J., and Lindauer, B.
\newblock 2013.
\newblock Bridging the gap: A pragmatic approach to generating insider threat
  data.
\newblock In {\em Proc. SPW},  98--104.

\bibitem[\protect\citeauthoryear{Hochreiter and
  Schmidhuber}{1997}]{hochreiter1997long}
Hochreiter, S., and Schmidhuber, J.
\newblock 1997.
\newblock Long short-term memory.
\newblock {\em Neural computation} 9(8):1735--1780.

\bibitem[\protect\citeauthoryear{Hutchins, Cloppert, and
  Amin}{2011}]{hutchins2011intelligence}
Hutchins, E.~M.; Cloppert, M.~J.; and Amin, R.~M.
\newblock 2011.
\newblock Intelligence-driven computer network defense informed by analysis of
  adversary campaigns and intrusion kill chains.
\newblock {\em Leading Issues in Information Warfare \& Security Research}
  1:80.

\bibitem[\protect\citeauthoryear{Kingma and Ba}{2014}]{kingma2014adam}
Kingma, D., and Ba, J.
\newblock 2014.
\newblock Adam: A method for stochastic optimization.
\newblock {\em arXiv preprint arXiv:1412.6980}.

\bibitem[\protect\citeauthoryear{Lindauer \bgroup et al\mbox.\egroup
  }{2014}]{lindauer2014generating}
Lindauer, B.; Glasser, J.; Rosen, M.; Wallnau, K.~C.; and ExactData, L.
\newblock 2014.
\newblock Generating test data for insider threat detectors.
\newblock {\em Journal of Wireless Mobile Networks, Ubiquitous Computing, and
  Dependable Applications} 5(2):80--94.

\bibitem[\protect\citeauthoryear{Liu, Ting, and Zhou}{2008}]{LiuTingZhou08}
Liu, F.~T.; Ting, K.~M.; and Zhou, Z.-H.
\newblock 2008.
\newblock Isolation forest.
\newblock In {\em Proc. ICDM}.

\bibitem[\protect\citeauthoryear{Malhotra \bgroup et al\mbox.\egroup
  }{2016}]{malhotra2016lstm}
Malhotra, P.; Ramakrishnan, A.; Anand, G.; Vig, L.; Agarwal, P.; and Shroff, G.
\newblock 2016.
\newblock {LSTM}-based encoder-decoder for multi-sensor anomaly detection.
\newblock {\em arXiv preprint arXiv:1607.00148}.

\bibitem[\protect\citeauthoryear{Marchi \bgroup et al\mbox.\egroup
  }{2015a}]{marchi2015novel}
Marchi, E.; Vesperini, F.; Eyben, F.; Squartini, S.; and Schuller, B.
\newblock 2015a.
\newblock A novel approach for automatic acoustic novelty detection using a
  denoising autoencoder with bidirectional {LSTM} neural networks.
\newblock In {\em Proc. ICASSP},  1996--2000.

\bibitem[\protect\citeauthoryear{Marchi \bgroup et al\mbox.\egroup
  }{2015b}]{marchi2015non}
Marchi, E.; Vesperini, F.; Weninger, F.; Eyben, F.; Squartini, S.; and
  Schuller, B.
\newblock 2015b.
\newblock Non-linear prediction with {LSTM} recurrent neural networks for
  acoustic novelty detection.
\newblock In {\em Proc. IJCNN},  1--7.

\bibitem[\protect\citeauthoryear{Markou and Singh}{2003}]{markou2003novelty}
Markou, M., and Singh, S.
\newblock 2003.
\newblock Novelty detection: a review—part 2:: neural network based
  approaches.
\newblock {\em Signal processing} 83(12):2499--2521.

\bibitem[\protect\citeauthoryear{Pedregosa \bgroup et al\mbox.\egroup
  }{2011}]{scikit-learn}
Pedregosa, F.; Varoquaux, G.; Gramfort, A.; Michel, V.; Thirion, B.; Grisel,
  O.; Blondel, M.; Prettenhofer, P.; Weiss, R.; Dubourg, V.; Vanderplas, J.;
  Passos, A.; Cournapeau, D.; Brucher, M.; Perrot, M.; and Duchesnay, E.
\newblock 2011.
\newblock Scikit-learn: Machine learning in {P}ython.
\newblock {\em Journal of Machine Learning Research} 12:2825--2830.

\bibitem[\protect\citeauthoryear{Ryan, Lin, and
  Miikkulainen}{1998}]{ryan1998intrusion}
Ryan, J.; Lin, M.-J.; and Miikkulainen, R.
\newblock 1998.
\newblock Intrusion detection with neural networks.
\newblock {\em Advances in neural information processing systems}  943--949.

\bibitem[\protect\citeauthoryear{Schölkopf \bgroup et al\mbox.\egroup
  }{2001}]{Scholkopfetal01}
Schölkopf, B.; J.~Platt~and, J. S.-T.; Smola, A.~J.; and Williamson, R.~C.
\newblock 2001.
\newblock Estimating the support of a high-dimensional distribution.
\newblock {\em Neural Computation} 13:1443--1471.

\bibitem[\protect\citeauthoryear{Shyu \bgroup et al\mbox.\egroup
  }{2003}]{ShyuChenetal03}
Shyu, M.-L.; Chen, S.-C.; Sarinnapakorn, K.; and Chang, L.
\newblock 2003.
\newblock A novel anomaly detection scheme based on principal component
  classifier.
\newblock In {\em Proc. ICDM}.

\bibitem[\protect\citeauthoryear{Sommer and Paxson}{2010}]{Sommer2010outside}
Sommer, R., and Paxson, V.
\newblock 2010.
\newblock Outside the closed world: On using machine learning for network
  intrusion detection.
\newblock In {\em Proc. Symposium on Security and Privacy}.

\bibitem[\protect\citeauthoryear{Veeramachaneni and
  Arnaldo}{2016}]{veeramachaneniai2}
Veeramachaneni, K., and Arnaldo, I.
\newblock 2016.
\newblock ${AI}^2$: Training a big data machine to defend.
\newblock In {\em Proc. HPSC and IDS}.

\end{thebibliography}
\bibliographystyle{aaai}

\end{document}